\documentclass{isprs} 
\usepackage{subfigure}
\usepackage{setspace}
\usepackage{geometry} 
\usepackage{epstopdf}
\usepackage[labelsep=period]{caption}  
\usepackage[british]{babel} 
\usepackage[hang]{footmisc}

\usepackage{graphicx}
\usepackage{color}

\usepackage{soul}
\usepackage{amssymb}
\usepackage{wrapfig,lipsum}

\begin{document}

\title{Landscape of Neural Architecture Search across sensors: how much do they differ ?}
\version{December 2020}


\author{Kalifou Ren\'e Traor\'e\textsuperscript{a, b}, Andr\'es Camero\textsuperscript{b, }, and Xiao Xiang Zhu\textsuperscript{a, b}}

\address
{
	\textsuperscript{a }Data Science in Earth Observation,
	Technical University of Munich\\	
	\textsuperscript{b }Remote Sensing Institute,
	German Aerospace Center (DLR)\\
	contact:\{kalifou.traore, andres.camerounzueta, xiaoxiang.zhu\}@dlr.de\\
}


\commission{III, }{YY} 
\workinggroup{III/6} 
\icwg{}   

\abstract{
With the rapid rise of \emph{neural architecture search}, the ability to understand 
its complexity from the perspective of a search algorithm is desirable.
Recently, Traoré et al. have proposed the framework of \emph{Fitness Landscape Footprint} 
to help describe and compare \emph{neural architecture search} problems. 
It attempts at describing why a search strategy might be successful, struggle or fail on a target task.
Our study leverages this methodology in the context of searching across sensors, including sensor data fusion.
In particular, we apply the \emph{Fitness Landscape Footprint} 
to the real-world image classification problem of So2Sat LCZ42, 
in order to identify the most beneficial sensor to our neural network hyper-parameter optimization problem.
From the perspective of distributions of fitness, 
our findings indicate a similar behaviour of the search space for all sensors: 
the longer the training time, the larger the overall fitness, and more flatness in the landscapes 
(less ruggedness and deviation).
Regarding sensors, the better the fitness they enable 
(Sentinel-2), 
the better the search trajectories (smoother, higher persistence).
Results also indicate very similar search behaviour for sensors 
that can be decently fitted by the search space (Sentinel-2 and fusion).
}
\keywords{AutoML, Neural Architecture Search, Fitness Landscape Analysis, Sensor Fusion, Remote Sensing}
\maketitle
\section{Introduction}



\emph{Neural architecture search} (NAS) is a rapidly growing area of \emph{machine learning} (ML) 
dedicated to automatically designing high performing \emph{deep learning} models.
Recent breakthroughs, such as \textit{differentiable search}, e.g., DARTS \cite{DARTS},
have enabled search at limited computing cost and time.
However, state-of-the-art methodologies still suffer from limited interpretability,
and current evaluation protocols do not always shed light on the contribution of individual components 
(i.e., search space, training pipeline) while reporting performances \cite{Yang2020NasEvalHard,Lindauer2020BestPractices}.

Recently, a \emph{fitness landscape analysis}-based (FLA) methodology was introduced: the \emph{Fitness Landscape Footprint} (a.k.a., \emph{footprint}) \cite{traore2021fitness}.
The \emph{footprint} attempts to describe 
why a search strategy may be successful, struggle or fail on a target application.
It also enables comparing search problems of variable configuration 
(i.e., different search space, fitness function, data, etc.).

Our study takes advantage of the \emph{footprint} to identify the most favorable sensor setting 
for NAS.
Particularly, we consider optimizing convolutional neural network (CNN) image classifiers on the search space defined by NASBench-101 \cite{ying2019nasbench101} for the real-world image classification problem So2Sat LCZ42~\cite{So2SatDataset}.
Our results show that disregard the sensor, the longer the training time, the better the performance (fitness) and the flatter the landscape (less ruggedness and deviation in fitness). 
Moreover, Sentinel-2 and fusion (Sentinel-1 and 2) tend to have more favorable search trajectories (smoother, higher persistence).
To the best of our knowledge, our study provides the first quantification 
and comparison of search behavior across sensors (including sensor fusion).

\begin{figure}[h]
\centering
    \includegraphics[width=0.5\textwidth]{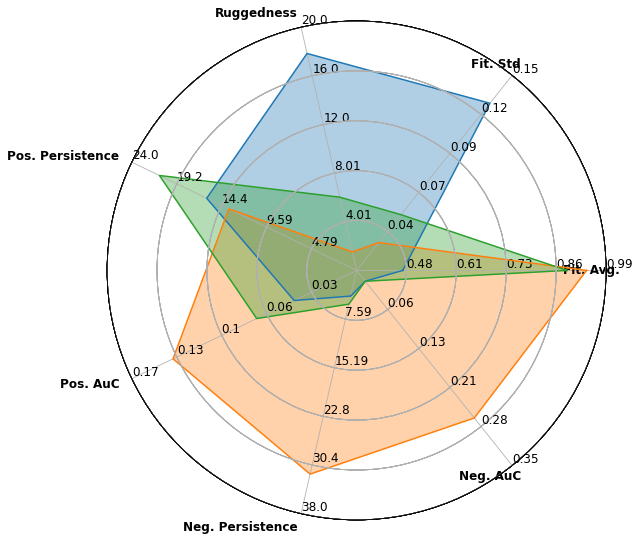}
    \caption{\emph{Fitness landscape footprint} of So2Sat LCZ42 for Sentinel-1 (blue), Sentinel-2 (yellow), and both sensors together (green), on CNN image classifiers NAS problem.} 
    \label{fig:footprnts}
\end{figure}

This article is structured as follows: 
Next section summarizes the related work and Section~\ref{sec:fitness-landscape} introduces the \emph{footprint}. 
Section~\ref{sec:footprint-for-fusion} proposes the methodology to study NAS problems. 
Sections \ref{sec:exp-setup} and~\ref{sec:results} presents the experimental settings and the results.
Section~\ref{sec:conclusion} outlines the conclusions and proposes future work.

\section{Related Work}\label{sec:related-work}

\emph{Computer vision} (CV) and \emph{Earth observation} (EO) are closely tied~\cite{ball2017comprehensive,zhu2017deep}.
Deep learning models in CV have helped tackle several specific EO use-cases such as 
scene classification~\cite{LiuResNet2019}, 
object detection~\cite{ZhangRCNN2019}, 
change detection~\cite{Mou2019RCNN_change_detection}, and 
semantic segmentation~\cite{YUAN2021Review_Semantic_Segm}, among others.
However, the specificity of sensors require domain-specific models~\cite{R3Net}.
In particular, the availability of several sensors to monitor areas 
has motivated the activity of multi-modal sensor fusion which remains challenging for current models~\cite{Hong2021MoreDiverseFusion}.
Moreover, as the design of new vision-based methodologies can be time-consuming  (trial and error),
EO could benefit from \emph{automated machine learning} (AutoML) algorithms.

In AutoML, NAS specializes in finding model configurations 
achieving optimal performances for a given dataset~\cite{elsken2019neural,ojha2017review}.
NAS methodologies have been proven to be powerful and efficient,
with strategies deriving from various families of optimization algorithm, e.g., 
\emph{differentiable search}~\cite{DARTS,traore2021esann}, 
\emph{Bayesian optimization}~\cite{camero2021bayesian}, \emph{meta-heuristic}-based approaches~\cite{stanley2002evolving,camero2020random,traore2021DataDrivenInit}.
However, in practice, the difficulty of a NAS problem is hard to estimate, because the complexity of its components, namely the
search space, search strategy, performance estimator and additional \emph{tricks}, is hard to quantify~\cite{elsken2019neural,Yang2020NasEvalHard}.

The fields of \emph{evolutionary computation}, \emph{optimization} and \emph{complex systems} have long studied optimization processes and provide us with tools to analyze their behavior.
In particular, FLA~\cite{pitzer2012comprehensive} aims at understanding 
and predicting performances of optimization algorithms.
Recently, Traoré et al. proposed the \emph{Fitness Landscape Footprint}~\cite{traore2021fitness}, 
a framework to characterize NAS problems from the perspective of a search algorithm. 
The following section introduces the \emph{footprint}.

\section{A Framework for Comparative Fitness Landscape Analysis}\label{sec:fitness-landscape}

Before describing the \emph{footprint}, it is important to define what a fitness landscape is.
Let~$S$ be the set of all possibles solutions of an optimization problem, i.e., the search space.
Let~$f$ be the fitness function, which attributes to each candidate solution~$x \in S$, 
a fitness measurement~$f(x) \in \mathbb{R}$.
Let~$N$ be a function providing a structure to the search space~$S$, the neighborhood relationship operator. 
Then, the fitness landscape~$\mathcal{L} = (S, f, N)$ consists of combining the three above, in order 
to provide respectively with a set of possibles solutions, a function to evaluate them and another to interconnect them.

Given this definition, we are interested in a better understanding of a NAS optimization process. 
The \emph{footprint}~\cite{traore2021fitness} serves this purpose 
of gaining insights into the process by describing its fitness landscape  
with a set of eight (8) metrics measuring aspects such as
the distribution of fitness, ruggedness of the landscape or persistence of fitness. 
A~\emph{footprint} includes: the mean and variance of fitness over~$S$, the ruggedness~$\tau$, 
an enumeration of local optima, the positive and negative persistence and their area under the curve ($AuC$).
The following paragraphs describe these metrics.

The \emph{fitness distance correlation} (FDC) is  
often interpreted a measure of the existence of 
search trajectories from randomly picked solutions to the known global optimum.
In practice, the FDC is not collected as a correlation score, 
but visualized as the distribution of fitness versus 
distance to the global optimum. 
It writes as~$FDC(f,x^*, S)=\{(d(x^*,y), f(y)) \mid \forall y \in S\}$, 
where~$S$ denotes the search space, 
$x^* \in S$ is the global optimum, 
~$d$ a distance function.   

The \emph{ruggedness} of the landscape also helps assessing the difficulty of the process tackled.
Let's consider a random walk~$RW$ in~$S$ of $N$ steps (models) and its corresponding fitness values.
The ruggedness~$\tau$ consists in the auto-correlation length over ~$RW$ :~$\tau=\frac{1}{\rho(1)}$, where ~$\rho(k=1)$ is the serial-correlation coefficient for consecutive lags. 

In \cite{traore2021fitness}, the authors propose the metric of \emph{persistence} characterizing the behavior of image classification models overtime. It measures the chances of solutions in the search space, 
to keep a rank~$N$ (top or bottom rank, based on fitness in test), as the training time grows.
This metric is complemented with its area under the curve ($AuC$), 
measuring the evolution of the persistence as $N$ grows.

Another way to characterize an optimization fitness landscape is to assess the existence of \emph{local optima}.
As some search algorithms might get stuck in such sub-optimal areas of~$S$, 
an enumeration~\cite{Hernando2012CardOptimaBenchmark} could help measure the difficulty of the search problem.

Last but not least, the~\emph{footprint} not only characterizes individual NAS landscapes, 
but also enables the comparison of a handful considering potential changes 
in either components~$S$, ~$f$, or~$N$.

\section{Assessing and comparing the landscapes of NAS for various sensors}\label{sec:footprint-for-fusion}

This study aims at investigating how the process of searching for neural architectures 
is affected by the type of sensor available as input.
More precisely, we seek to identify to what extent 
does searching with a given sensor 
differs from searching with another one.
In particular, we consider the case of a fixed search space, 
training pipeline (hyperparameters, duration, etc.)
and evaluation protocol (fitness function). 
In practice, we propose to tackle these questions by conducting 
a comparative landscape study using the \emph{footprint} (Section~\ref{sec:fitness-landscape}).

Let~$\Sigma=\{s_i, s_j, s_i + s_j\}$ be the set of sensors available in our ML task.
Let~$S$ be a search space of CNN image classifiers, each represented by a unique binary vector.
Considering this representation, we choose a neighborhood operator~$N(x)$ assigning to each solution of the search space, all the configurations that are one (1) 
\textit{hamming distance} away from it. This operator~$N(x)$ writes as follows:~$N(x) = \{ y \in S \mid d_{hamming(x, y)} = 1\}$, 
where~$d_{hamming(x, y)}$ is the \textit{hamming distance} between two solutions~$(x,y) \in S^2$.
Additionally, we use as fitness function~$f_{s_{i}}$, the measurement of accuracy in test 
after a training budget of~$b_t$, on an input sensor~$s_i$.

Since we have access to various sensors~$s \in \Sigma$ for our ML task, 
the fitness landscapes obtained write as follows:
~$\mathcal{L}_{s_i} = (S, f_{s_i}, N)$, ~$\mathcal{L}_{s_j} = (S, f_{s_j}, N)$, ~$\mathcal{L}_{s_k} = (S, f_{s_k}, N)$ for the sensor settings $(s_i, s_j, s_k) \in \Sigma^3$.
In particular, we consider the case of input level sensor fusion as:~$ s_k = s_i + s_j$.
Besides, as noted above, the aim is a comparative study so we fix the search space~$S$ and the neighborhood operator~$N$ across all settings.

\section{Experimental Setup}\label{sec:exp-setup}
This section introduces the NASBench-101 database, as well as a custom representation used to encode its solutions.
Then, the So2Sat LCZ42 dataset used to evaluated solutions, followed by details on the evaluation protocol.

\subsection{NASBench-101}

NASBench-101~\cite{ying2019nasbench101} is a database containing a large pool of neural networks 
and their evaluations on the image classification dataset of CIFAR-10.
It aims at providing an exhaustive fitness measurement for all configurations (N=453k) 
in a search space of CNN image classifiers. 
This search space defines a model configuration as an image classification backbone 
with a head, body and tail.
Its body consists of repeating three identical 'block' structures alternated with down-sampling modules.
Regarding each block, it consist in a sequence of identical and elementary feed-forward units called cells.
Each cell is represented by a directed DAG with a maximum number of nodes ($V \leq 7$), 
maximum number of edges ($E \leq 9$) and a fixed listed of three (3) operators (Max-pool 3x3, Convolutional layer 1x1 and 3x3) 
labelling each node. Therefore, a solution of the search space is identified by a cell, 
encoded in practice by both an adjacency matrix of variable size (upper triangular), and its list of operators.
Moreover, The head of the model is a 3 x 3 convolution with 128 output channels, while the tail is a dense softmax layer.

\subsection{Custom feature representation}

In our experiments, 
we construct a custom representation to enable solutions of the search space to be identified by a single vector.
First, for our representation of the DAG, we do not label nodes.
Instead, for the five (5) intermediate nodes out of seven (7) (one for IN  and OUT), 
we account for the fact that each could be one of three (3) operators.
Thus, the DAG contains exactly $N_{nodes}=(1+5*3+1) = 17$ nodes.  
The new adjacency matrix is therefore of fixed length, i.e $L=17*17$ and non upper-triangular.
Finally, we flatten the adjacency matrix to obtain a binary vector as identifier.

Regarding the sampling of solutions $x \in S$, we use the Latin Hypercube Sampling (LHS) for ensure fair data collection.
Because of a higher complexity of LHS on the large binary representation, we perform it on the intermediate 
representation as a joint sampling of the original matrix and list of operations.

\subsection{So2Sat LCZ42}

\begin{figure}[h]
    \centering
    \includegraphics[width=\columnwidth]{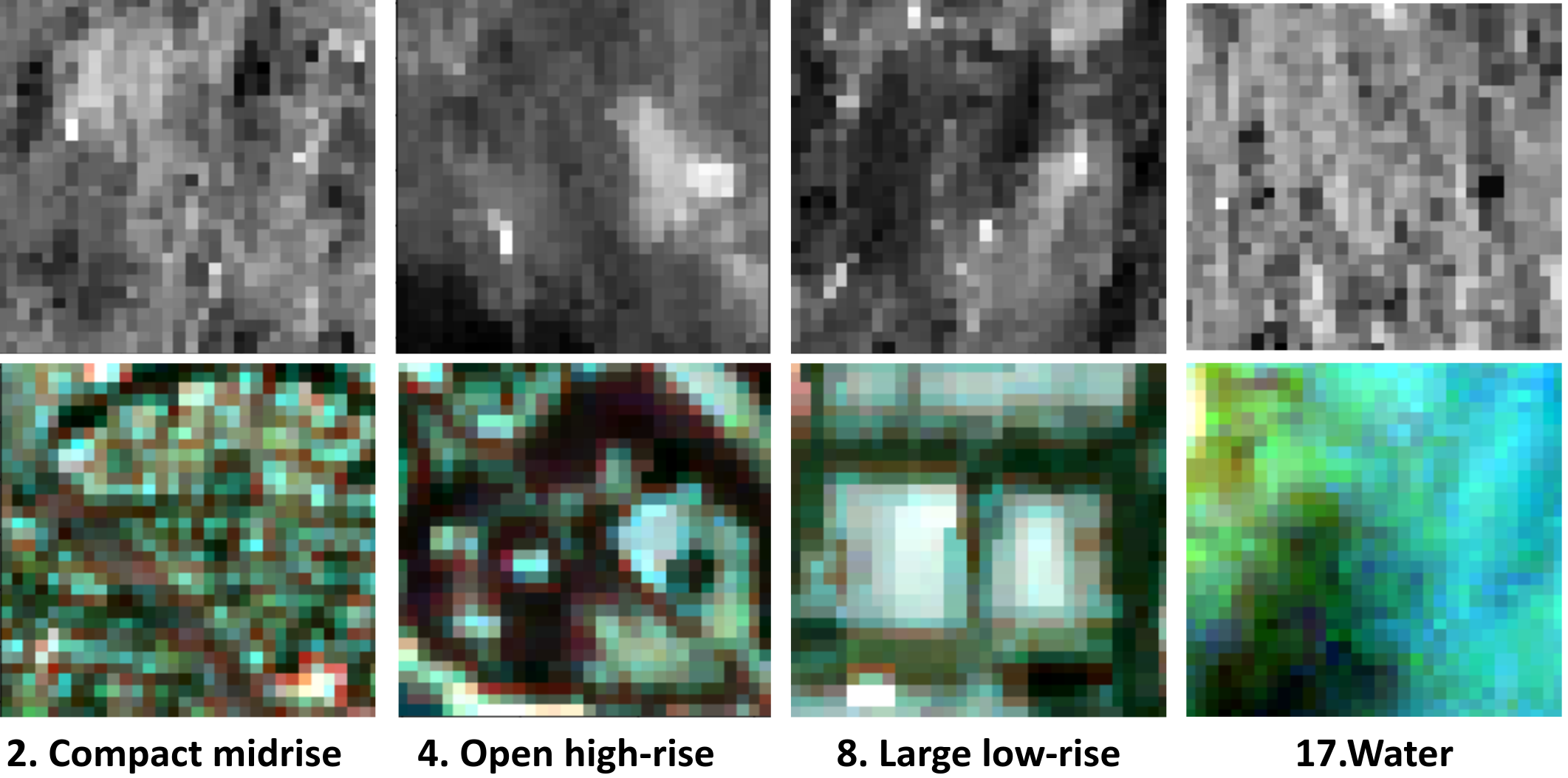}
    \caption{So2Sat LCZ42 samples. The top row contains SAR patches (Sentinel-1), followed by the associated Multi-spectral patches (Sentinel-2) in the bottom row.}
    \label{fig:lcz42-dataset-visual}
\end{figure}

The So2Sat LCZ42~\cite{So2SatDataset} is a dataset of satellite imagery covering over forty-two (42) cities around the five (5) continents.
It provides with co-registered image patches from Sentinel-1 and Sentinel-2 sensors, 
each attributed with a single label.
This label associates an image to a class out of seventeen (17) possibilities, 
all representing the diverse Local Climate Zones (LCZ) around the globe.
The framework of LCZ proposes a generic way to describe the morphology 
of land use around the world in both urban and non-urban natural sites. 
The classes are the following: Compact high-rise (1), Compact mid-rise (2), Compact low-rise (3), Open high-rise (4), 
Open mid-rise (5), Open low-rise (6), Lightweight low-rise (7), Large low-rise (8), Sparsely built (9), and Heavy industry (10), 
Dense trees (11), Scattered tree (12), Bush, scrub (13), Low plants (14), Bare rock or paved (15), Bare soil or sand (16), 
and Water (17). 
Figure~\ref{fig:lcz42-dataset-visual} displays four (4) pairs of Sentinel-1 and Sentinel-2 image patches respectively from classes 
2, 4, 8 and 17.

The dataset comprises train ($N_t=352,366$), validation ($N_v=24,119$), and test ($N_{test}=24,188$) samples.
The training and validation samples originate from the same set of forty-two (42) cites, 
while those from the test set were collected in ten (10) additional cities.

\subsection{Evaluation protocol}
For the purpose of training and evaluating on the same data distribution, 
we use a custom setting consisting of training and testing sets made by randomly sampling respectively, 
$80\%$ and $20\%$ of the image patches of the original training-set. 
Moreover, as in \cite{traore2021fitness} we speed-up the training procedure by only considering $P=35\%$ of samples in the training set.

Additionally, we use the same search space~$S$ for all sensor settings.
In particular, we do not adapt the sampled models to use multiple sensors as input, 
instead we do stack the data at an input level.
We trained ~$N=100$ randomly sampled models, once. After inspection and quality control, 
there remain 100, 88 and 75 samples for Sentinel-1, 2 and both sensors.
The fitness is assessed in test using the Kappa-Cohen metric.

\section{Results}\label{sec:results}

The following section presents results of comparison of search landscapes for various input sensors.
First, we provide an analysis of distributions of fitness.
Then, we show results of fitness distance correlation, followed by 
an analysis of random walks, as well as measurements of fitness persistence.
Last but not least, we compare the \emph{footprint} of the sensors.

\subsection{Density of Fitness}\label{subsec:dos}

\begin{figure*}[ht]
\centering
    \includegraphics[width=0.275\textwidth]{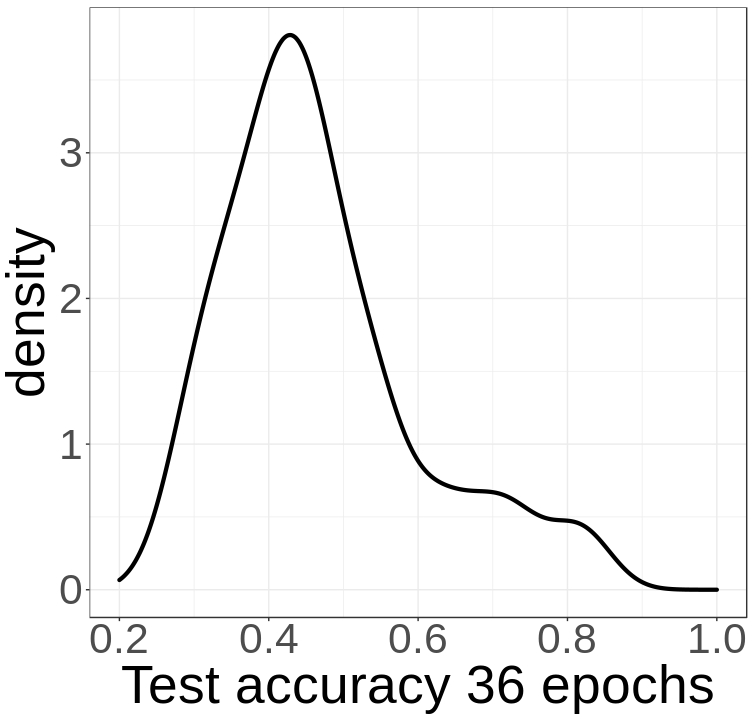}
    \includegraphics[width=0.275\textwidth]{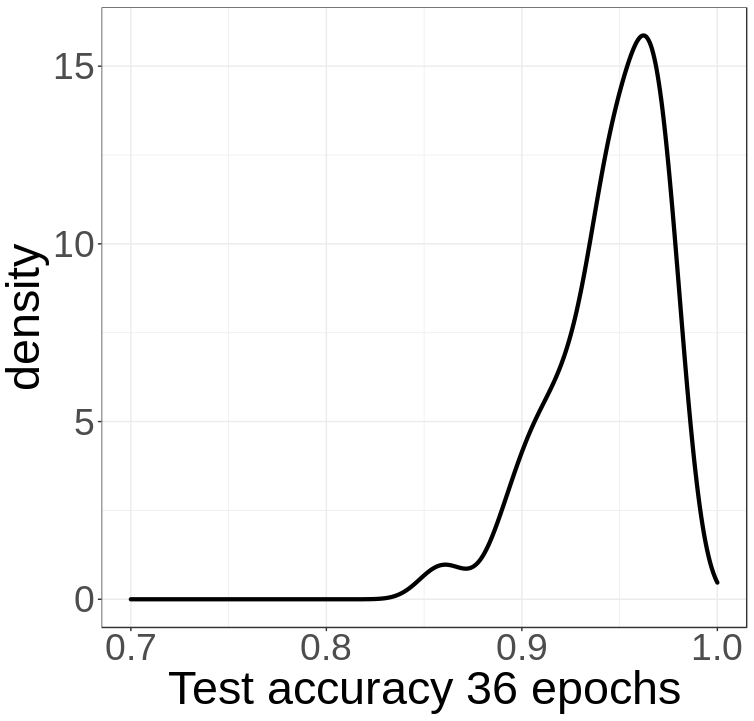}
    \includegraphics[width=0.275\textwidth]{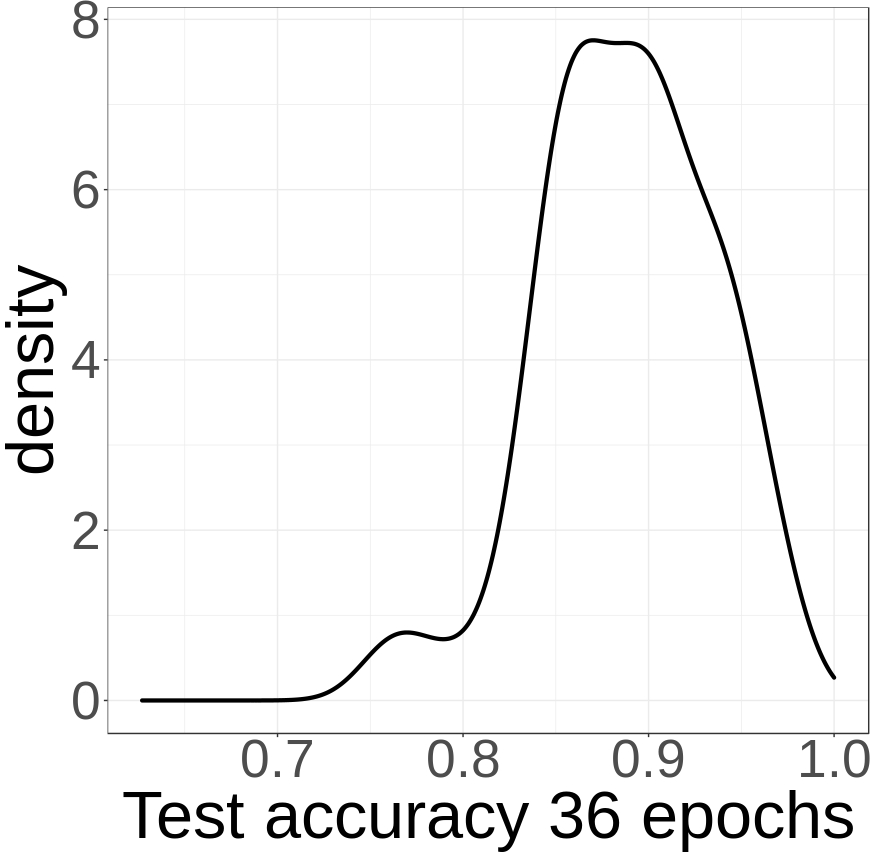}
    \caption{PDF of fitness after 36 epochs of training. 
    From left to right are results using Sentinel-1, Sentinel-2 and both sensors as input.
    } 
    \label{fig:pdf-36}
\end{figure*}

\begin{figure*}[ht]
\centering
    \includegraphics[width=0.275\textwidth]{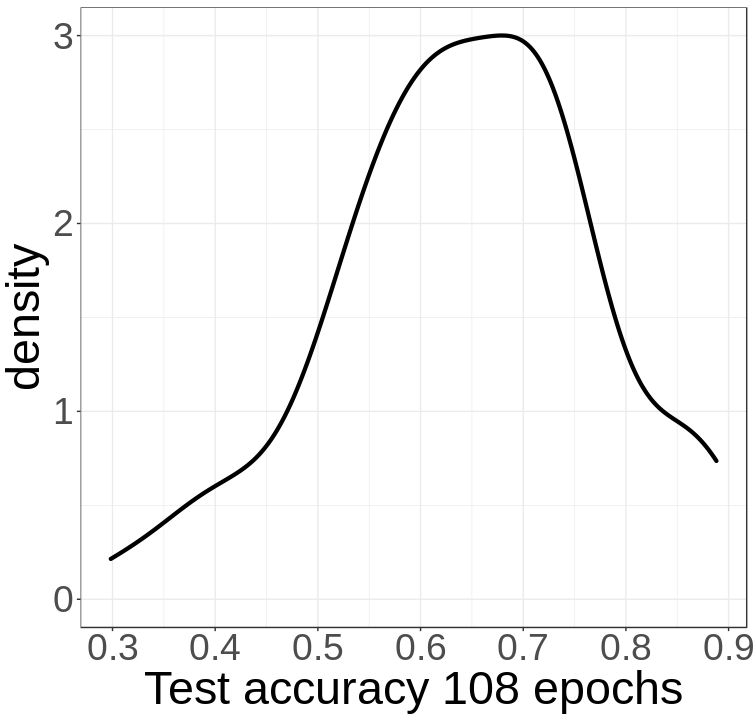}
    \includegraphics[width=0.275\textwidth]{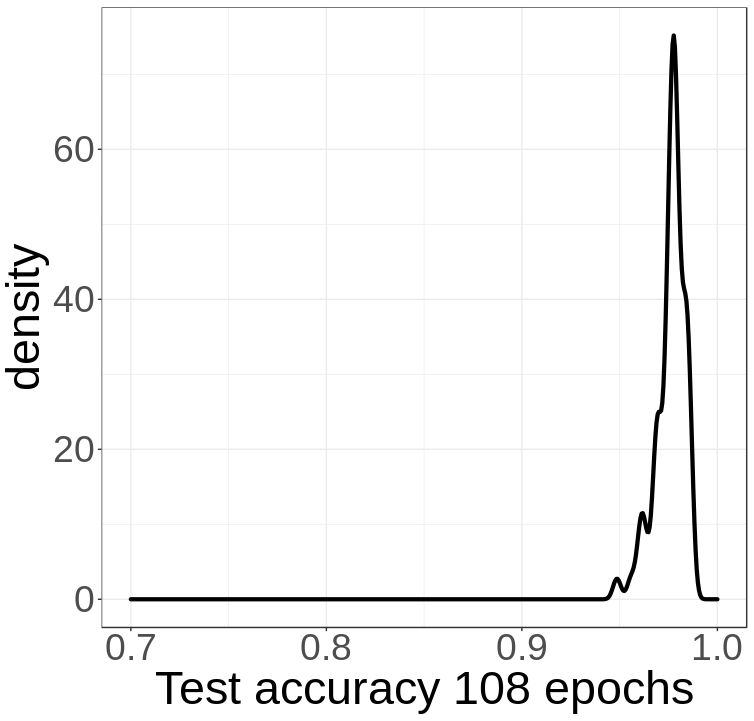}
    \includegraphics[width=0.275\textwidth]{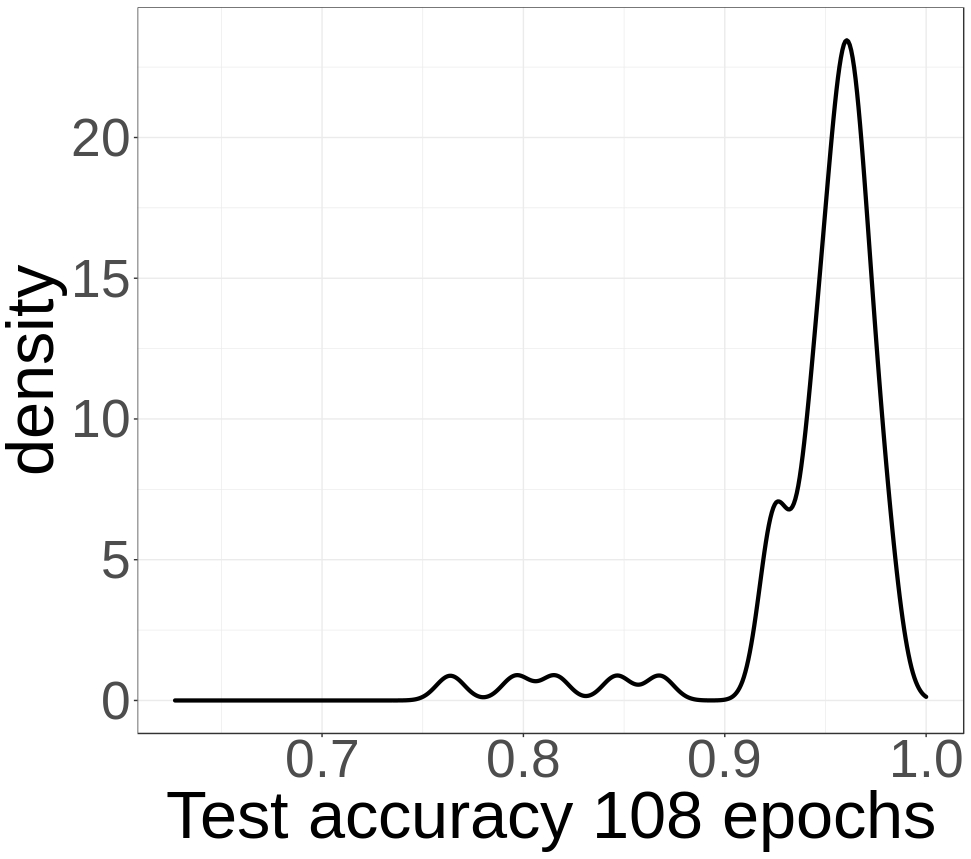}
    \caption{PDF of fitness after 108 epochs of training. 
    From left to right are results using Sentinel-1, Sentinel-2 and both sensors as input.
    } 
    \label{fig:pdf-108e}
\end{figure*}

First, we assess the ability of the search space in fitting the task with each sensor. 
Figure~\ref{fig:pdf-36} and ~\ref{fig:pdf-108e} display the probability density function (PDF) of fitness,
respectively after 36 and 108 epochs of training. 
The first, second and third columns are, respectively, for using Sentinel-1, Sentinel-2 or both sensors as input.

We first take a look at the PDFs after 36 epochs of training.
When using Sentinel-1, the distribution of fitness is wide and centered around low values ($\mu=0.47, \sigma=0.13$). 
Sentinel-2 enables the distribution to improve by reaching a higher average and being more narrow ($\mu=0.94, \sigma=0.03$).
Using both sensors slightly worsens the fitness, providing with a lower mean and larger deviation ($\mu=0.89, \sigma=0.05$).

Next, we look at the PDFs after 108 epochs of training.
Overall, the task is better handled in all sensor configurations.
Using Sentinel-1, the distribution improves by $17$ percentage points in mean fitness ($\mu=0.64, \sigma=0.13$).
In the case of Sentinel-2, most models fit well the data as the mean fitness increases and the deviation decreases ($\mu=0.97, \sigma=0.01$).
We observe similar results when using both sensors ($\mu=0.94, \sigma=0.04$).


\begin{figure*}[ht]
\centering
    \includegraphics[width=0.275\textwidth]{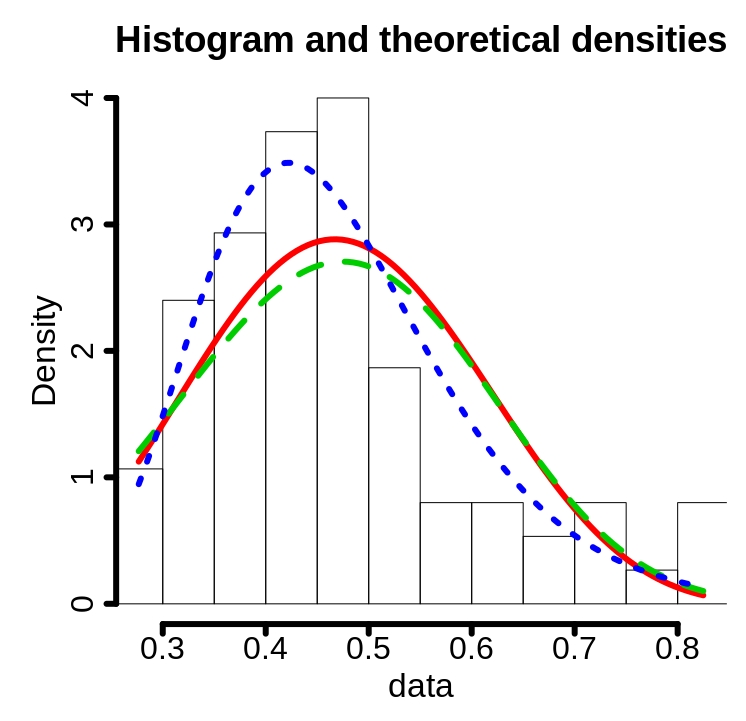}
    \includegraphics[width=0.275\textwidth]{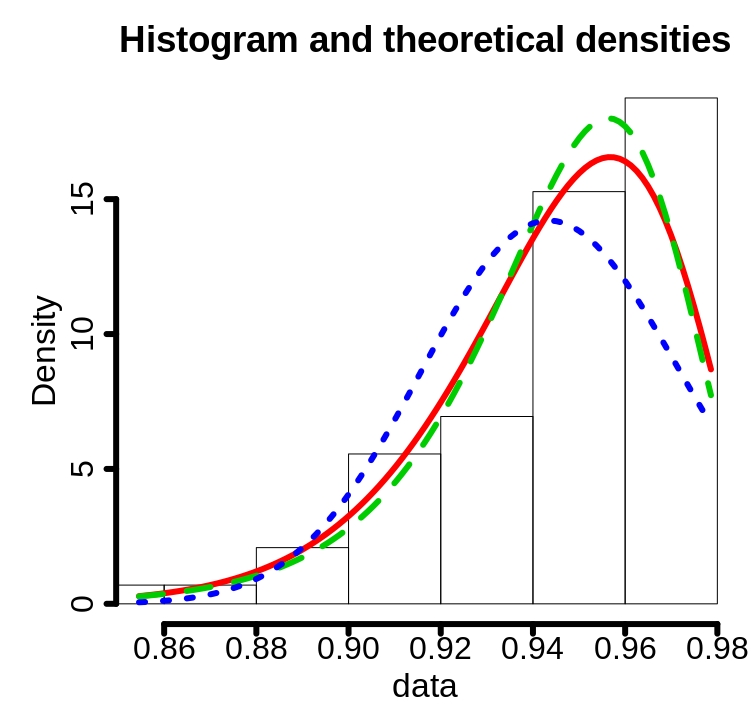}
    \includegraphics[width=0.275\textwidth]{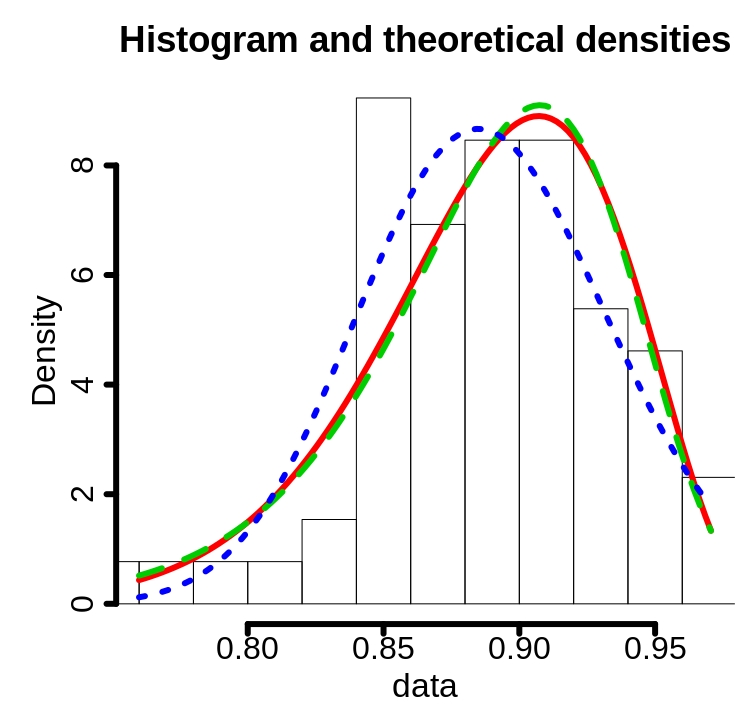}
    \caption{Fitting PDF to fitness with various theoretical functions after 36 epochs of training. 
    From left to right are results using Sentinel-1, Sentinel-2 and both sensors. 
    The red, blue and green colors correspond to Beta, Log-normal and Weibull distributions, respectively.
    } 
    \label{fig:pdf-fitting-36e}
\end{figure*} 


\begin{figure*}[ht]
\centering
    \includegraphics[width=0.275\textwidth]{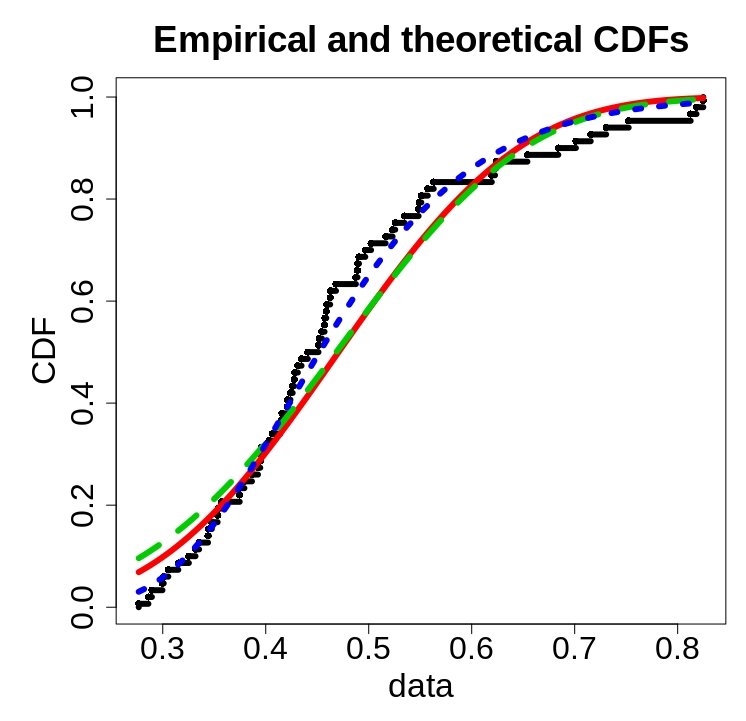}
    \includegraphics[width=0.275\textwidth]{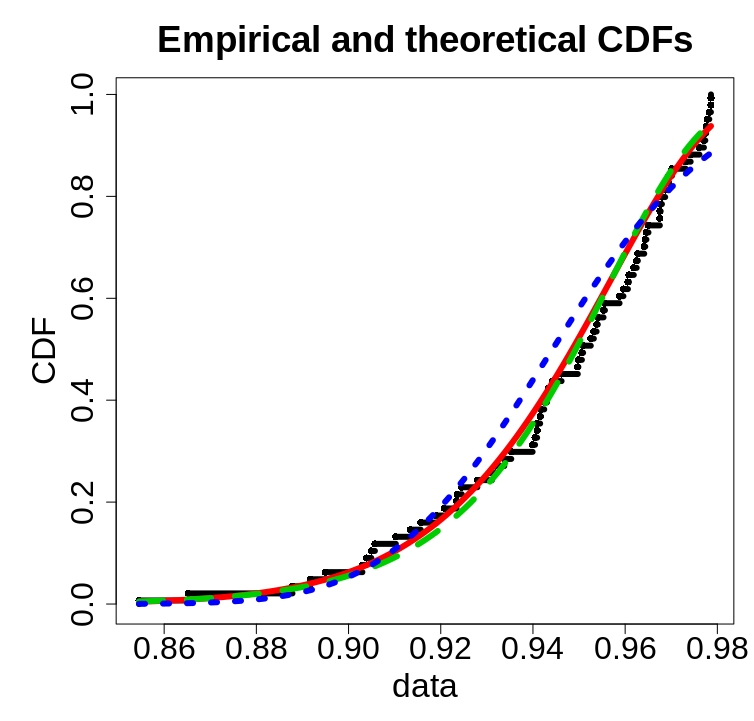}
    \includegraphics[width=0.275\textwidth]{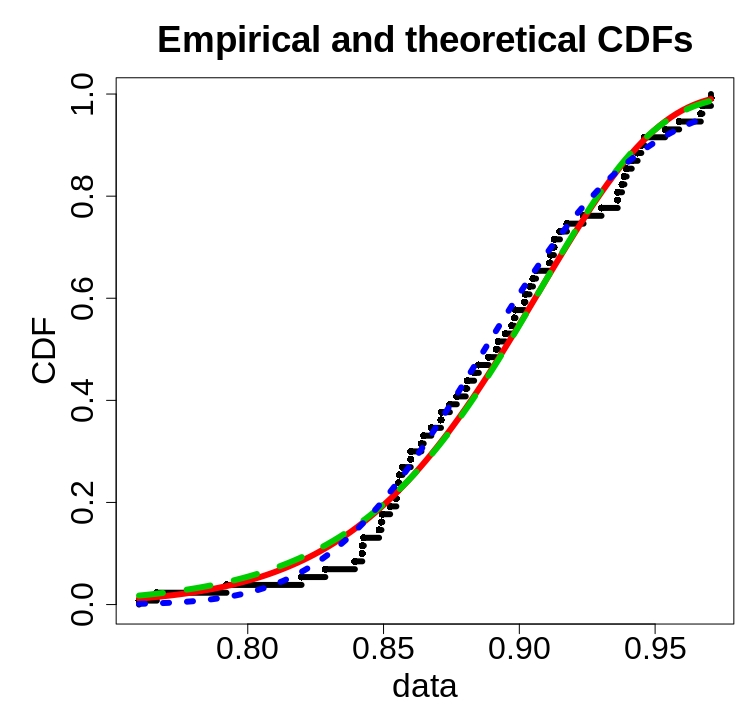}
    \caption{ Cumulative PDFs of fitness versus theoretical density functions, after 36 epochs of training.  
    The plot uses the same color and sensor convention as Figure~\ref{fig:pdf-fitting-36e}.
    } 
   \label{fig:cdf-36e}
\end{figure*} 


Besides, we seek to identify for each sensor, 
if the behavior of the search space follows a specific theoretical distribution. 
For this, we consider the more challenging scenario 
of selecting models after only 36 epochs of training.

Figure~\ref{fig:pdf-fitting-36e} and ~\ref{fig:cdf-36e} show results of fitting empirical distributions of fitness,
with various theoretical distributions.
Figure~\ref{fig:pdf-fitting-36e} and ~\ref{fig:cdf-36e} show, respectively, PDFs and  cumulative density functions (CDF), 
all fitted with the Beta (red), Weibull (green) and Log-normal (blue) distributions.
Similarly, the first, second and third columns are respectively, 
for using Sentinel-1, Sentinel-2 or both sensors as input.
To complement the plots, Table~\ref{tab:pdf-fitting-error-s1},~\ref{tab:pdf-fitting-error-s2} and~\ref{tab:pdf-fitting-error-s12} provide with the respective fitting errors.

\begin{table}[h!]
 \centering 
\scriptsize
\begin{tabular}{|l|l|l|l|l|}\hline 
\textit{Error Metric / Function} &\textit{Beta} & \textit{Weibull} & \textit{Log-normal}\\
\hline
Likelihood & 46.28   & 44.52 & \textbf{53.63} \\ 
\hline
AIC & -88.56  & -85.04  &  \textbf{-103.26} \\  
\hline
BIC &-83.92   & -80.41  &  \textbf{-98.62}\\    
\hline
\end{tabular}
\caption{PDF fitting error for Sentinel-1 } 
\label{tab:pdf-fitting-error-s1}
\end{table}

\vspace{0.2cm}

\begin{table}[h!]
 \centering 
\scriptsize
\begin{tabular}{|l|l|l|l|l|}\hline 
\textit{Error Metric / Function} &\textit{Beta} & \textit{Weibull} & \textit{Log-normal}\\
\hline
Likelihood & \textbf{165.18}   & 164.70 & 155.05 \\ 
\hline
AIC & \textbf{-326.36}  & -325.40  &  -306.10 \\  
\hline
BIC & \textbf{-321.80}   & -320.84  &  -301.54\\   
\hline
\end{tabular}
\caption{PDF fitting error for Sentinel-2 } 
\label{tab:pdf-fitting-error-s2}
\end{table}

\vspace{0.2cm}

\begin{table}[h!]
 \centering 
\scriptsize
\begin{tabular}{|l|l|l|l|l|}\hline 
\textit{Error Metric / Function} &\textit{Beta} & \textit{Weibull} & \textit{Log-normal}\\
\hline
Likelihood & \textbf{109.72}  & 109.04 & 107.78\\ 
\hline
AIC & \textbf{-215.44}  & -214.08  &  -211.57 \\  
\hline
BIC & \textbf{-211.09}   & -209.74  &  -207.22\\    
\hline
\end{tabular}
\caption{PDF fitting error for both sensors as input } 
\label{tab:pdf-fitting-error-s12}
\end{table}

Overall, the empirical distributions are closely fitted with the selected theoretical distributions.
For Sentinel-1, the best candidate is the Log-normal with the largest fitting likelihood, and lowest AIC and BIC error scores. When using Sentinel-2 or both sensors, Beta matches the best the empirical distributions. 

To summarize, 
the capacity of the search space in fitting the task with the Sentinel-1 sensor appears limited from the PDF perspective.
Indeed, despite longer training time (108 epochs) the fitness distribution remains far worse than using Sentinel-2.
Using Sentinel-2 only, the task can be fitted well enough in particular given long training time.
Combining Sentinel-1 to Sentinel-2 worsens the distribution of fitness (lower mean, larger deviation).
Therefore, there is no tangible benefits in fitness, from sensor fusion using the current search space.
Moreover, results indicate the feasibility in modeling the empirical distributions of fitness for each sensor.

\subsection{Fitness Distance Correlation}\label{subsec:fdc}

\begin{figure*}[ht]
\centering
    \includegraphics[width=0.3\textwidth]{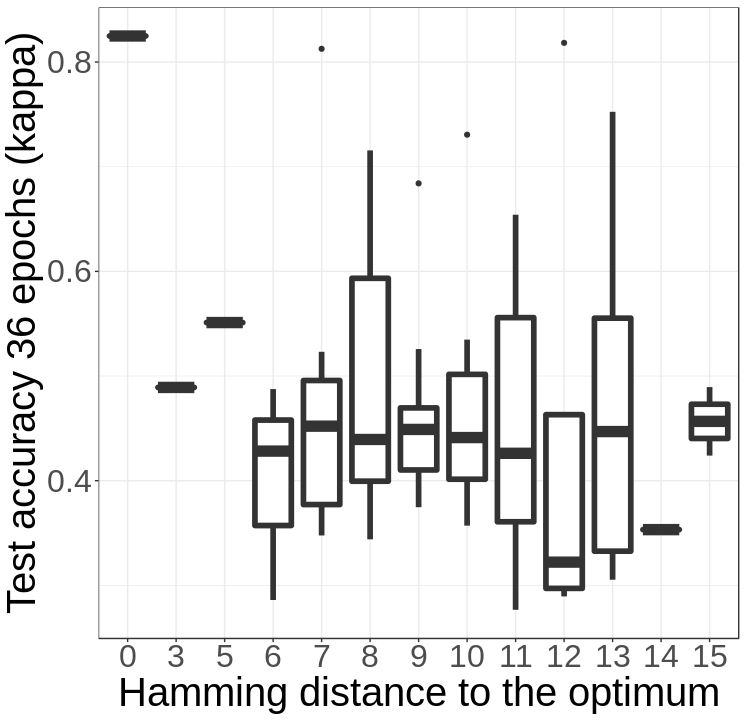}
    \includegraphics[width=0.3\textwidth]{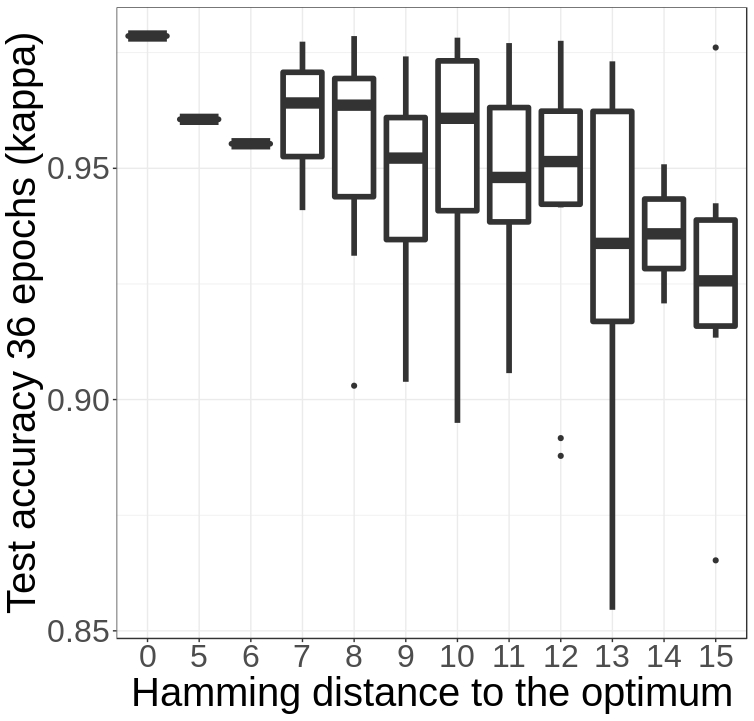}
    \includegraphics[width=0.3\textwidth]{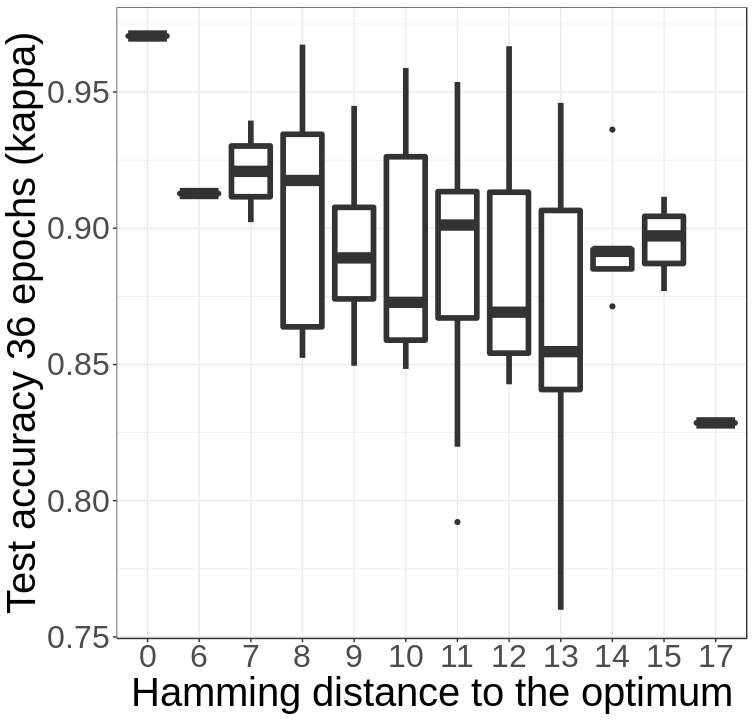}
    \caption{Fitness distance correlation after 36 epochs of training.
    From left to right are shown results using Sentinel-1, Sentinel-2 and both sensors. 
    } 
    \label{fig:profiles-36}
\end{figure*}

\begin{figure*}[ht]
\centering
    \includegraphics[width=0.275\textwidth]{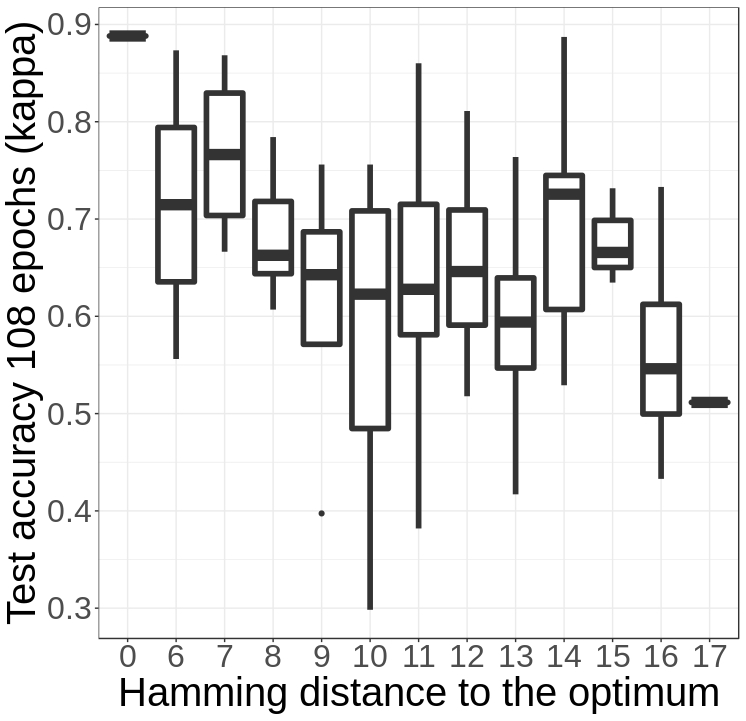}
    \includegraphics[width=0.275\textwidth]{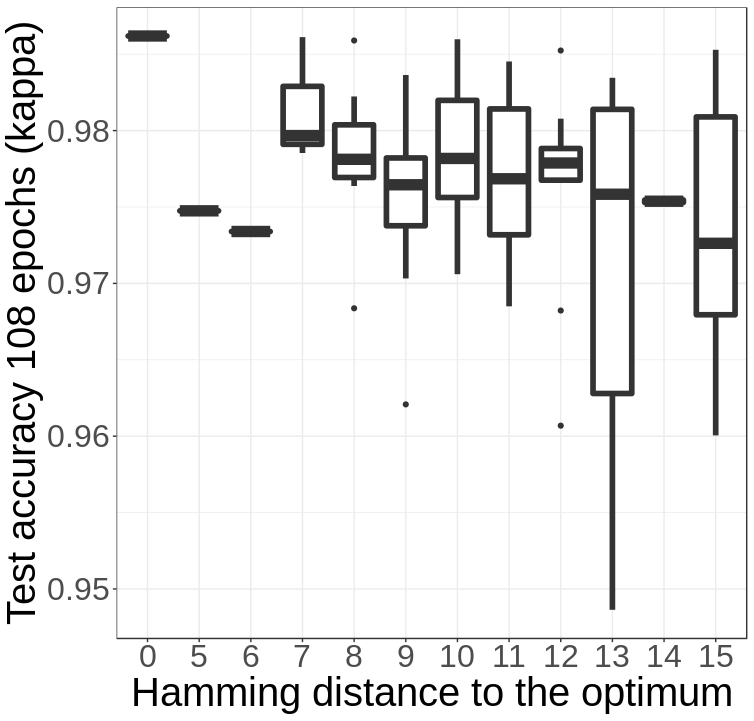}
    \includegraphics[width=0.275\textwidth]{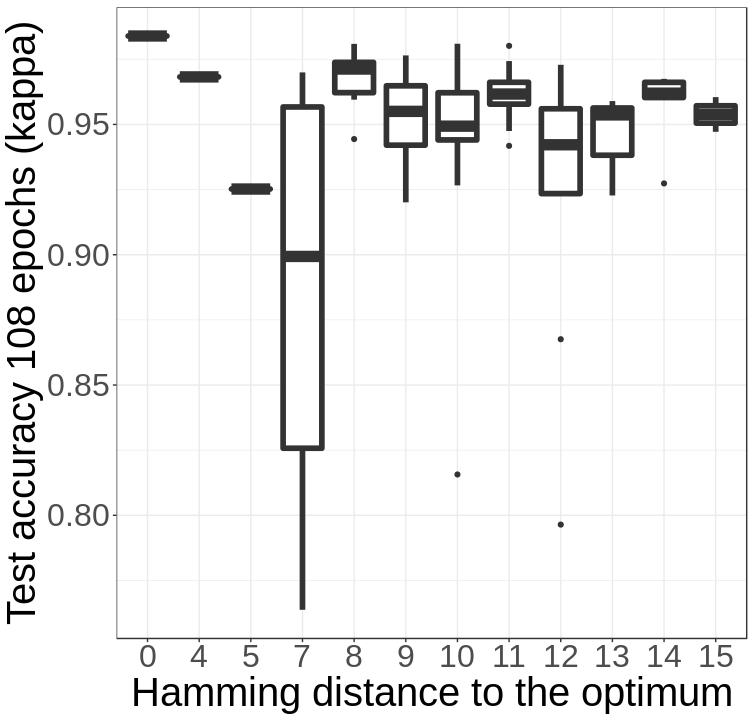}
    \caption{Fitness distance correlation after 108 epochs of training.
    From left to right are shown results using Sentinel-1, Sentinel-2 and both sensors. 
    } 
    \label{fig:profiles-108}
\end{figure*}

Next, we analyse the fitness landscape for the various sensor configurations.
Figure~\ref{fig:profiles-36} and \ref{fig:profiles-108} show results of FDC for the three (3) input sensor settings.
The layout of the plots follows the convention of Figure~\ref{fig:pdf-36}.

First, we consider the FDC after 36 epochs of training (see Figure~\ref{fig:profiles-36}).
Overall, for all sensors, we observe that the respective landscapes are rather rough.
Indeed, the distribution of fitness per hamming distance to the optimum are relatively wide.
For instance, when using Sentinel-1 we notice that solutions at the \textit{Hamming-distances} $d_{hamming}=\{8, 11, 13\}$ display up to 35\% percentage point in fitness difference.
Also, We notice a landscape around low fitness values (c.a 47\%).
Using Sentinel-2 provides with a landscape centered at much higher values (c.a 94\%).
We also notice a consistent increase in fitness, 
as the hamming distance to the optimum decreases.
Similarly, the landscape associated with using both sensors is of high fitness. 
However, the slope of gained fitness per travelled distance to the optimum worsens (less consistent), compared to the one obtained with Sentinel-2 as input.

Then, we consider the FDC after 108 epochs of training (see Figure~\ref{fig:profiles-108}).
Overall, the landscape tends to be more flat and with an increased fitness.
In particular for Sentinel-2 or both sensors as input, 
the flatness is indicated by more narrow distribution of fitness 
at the various distances to the optimum.
This also is complemented by potential search trajectories 
that have little improvements in fitness per travelled distance to the global optimum. 
Using both sensors brings us a similar behaviors, 
except for the existence of a set of models providing poorer fitness values, 
all located at $d_{hamming}=7$ from the optimum.
The case of Sentinel-1 is rather odd as there appears to be a favorable (negative) slope, as if the landscape had not converged.

To summarize, the fitness landscape of So2Sat LCZ42 is rougher when training is limited (36 epochs), 
and flatter towards higher fitness when training long enough (108 epochs) solutions in the search space.
As observed when analyzing distributions of fitness (Figure~\ref{fig:pdf-36}, \ref{fig:pdf-108e}), 
this NAS problem benefits better from using Sentinel-2 as input, 
with improvements in slope and overall fitness in its landscape.
Therefore, these results complement the analysis of PDFs 
by showing benefits in performances, this time from the perspective of potential NAS algorithm trajectories. It also shows that with the current search space, the search behaviour is worse when using both sensors as input.

\subsection{Random Walk Analysis}\label{subsec:random-walks}

Furthermore, we investigate the behavior 
of local search-based algorithm depending on the input sensor.
More precisely, this is done by analyzing random walks.

\begin{figure}
\centering
    \includegraphics[width=0.45\textwidth]{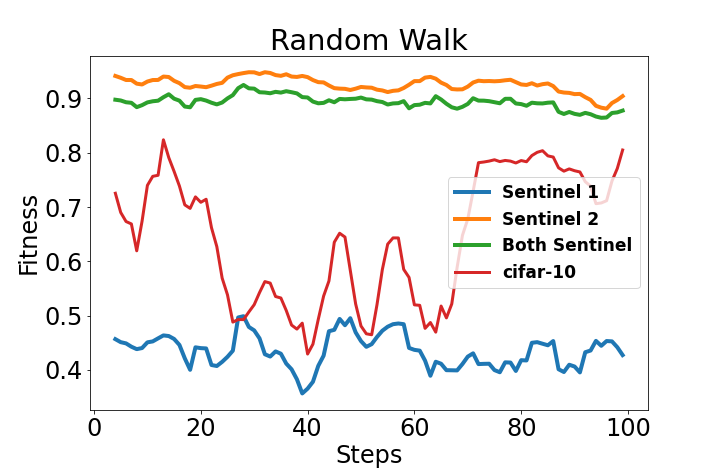}
    \caption{A random walk evaluated with the four (4) different sensor settings.  
    } 
    \label{fig:random-walks}
\end{figure}

Figure~\ref{fig:random-walks} displays the route of a random walk 
evaluated on for four (4) different sensor settings.
The walk itself consists of one hundred (100) steps in the search space.
At each step, the selected model is evaluated after being trained for 36 epochs. 
The blue, yellow, green and red curves are, respectively, 
for evaluating the fitness with Sentinel-1, Sentinel-2, both sensors, or CIFAR-10 as input. 
All curves were smoothed with a moving average of five (5) steps.
We also consider CIFAR-10, since its fitness evaluations were freely 
available (ground-truth in NAS-Bench-101) and could serve as reference for comparison, and trouble-shooting.

The evaluation of the walk on Sentinel-1 provides with the lowest overall fitness (~$\mu=0.44$) and the most rugged route (~$\tau=17.84$).
On the other hand, using Sentinel-2 or both sensors together, provides with more smooth paths, at much higher values. Indeed, the respective averages of fitness are ~$\mu=0.93$ and ~$\mu=0.89$. 
The ruggedness values are ~$\tau=1.54$ and ~$\tau=6.03$. 
Also , the curvature of both routes \textit{visually} look alike.
Regarding CIFAR-10, we observe intermediate fitness and ruggedness (~$\mu=0.64$ and ~$\tau=1.56$).
The relatively large amplitude in fitness and similar curvature, despite lower ruggedness, makes its route look more similar to the one evaluated with Sentinel-1.

As observed when analyzing FDCs (Section~\ref{subsec:fdc}),
Sentinel-1 provides with poorer trajectories (lower fitness, larger ruggedness), 
suggesting either a sensor being unsuitable for the task 
or a search space~$S$ not suitable for the sensor.
Similar results (curvature, lower fitness) obtained for ground-truth evaluations on CIFAR-10 suggest 
that a higher ruggedness seems to associate with harder tasks and lower convergence of models in a random walk route.

To summarize, the use of either Sentinel-2 or both sensors after only 36 epochs
enables a NAS route to be of higher smoothness and fitness.

\subsection{Persistence}\label{subsec:persistence}

Next, we study the behaviour of solutions in the search space, 
from the perspective of persistence in their ranking.

\begin{figure}[h]
\centering
    \includegraphics[width=0.4\textwidth]{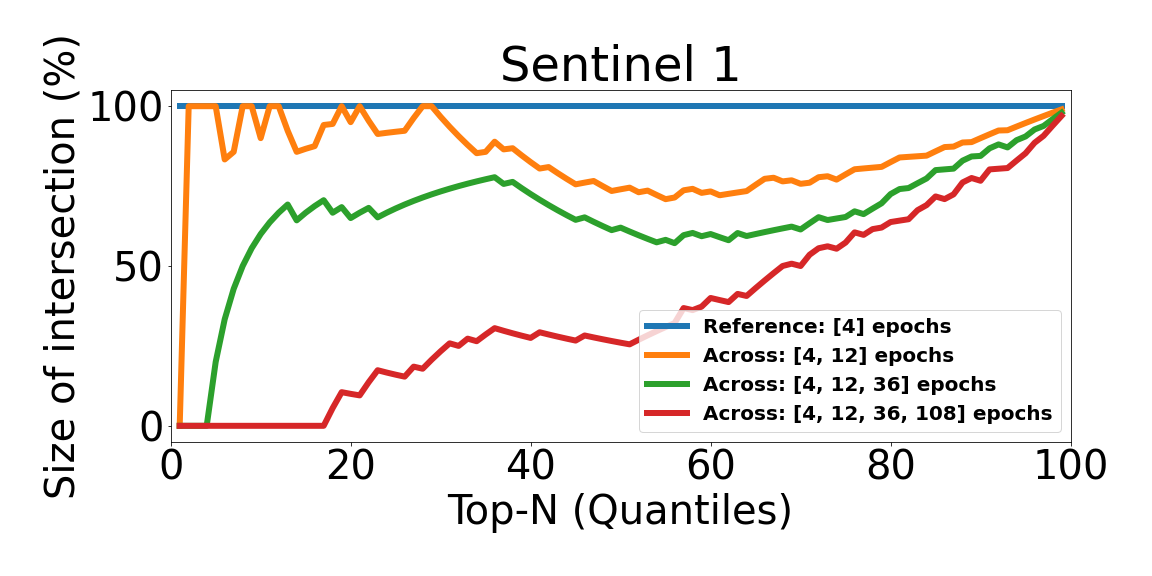}
    \qquad
    \includegraphics[width=0.4\textwidth]{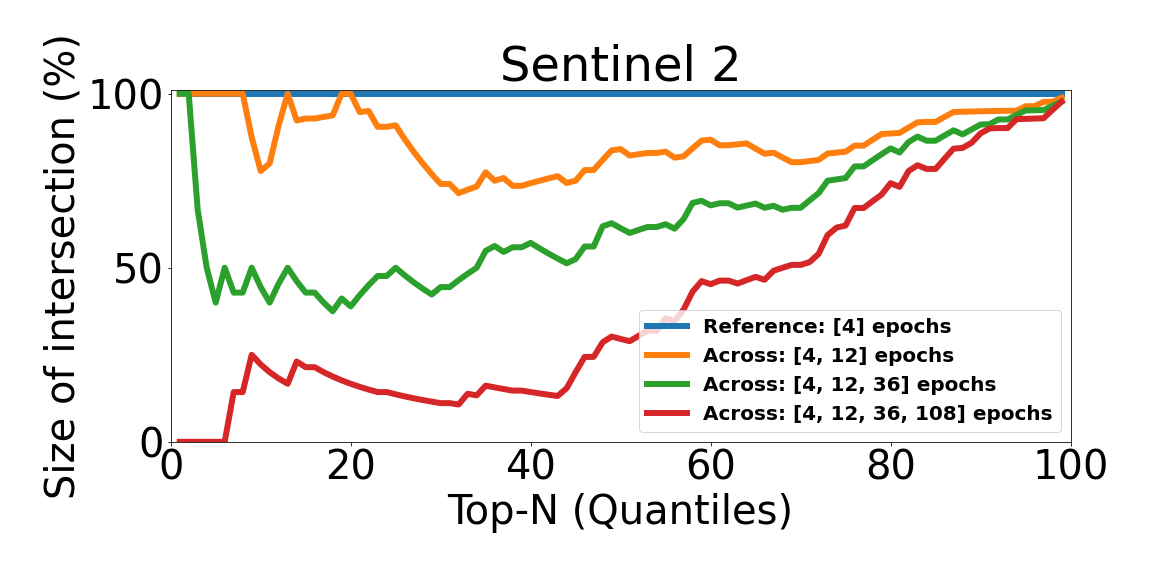}
    \qquad
    \includegraphics[width=0.4\textwidth]{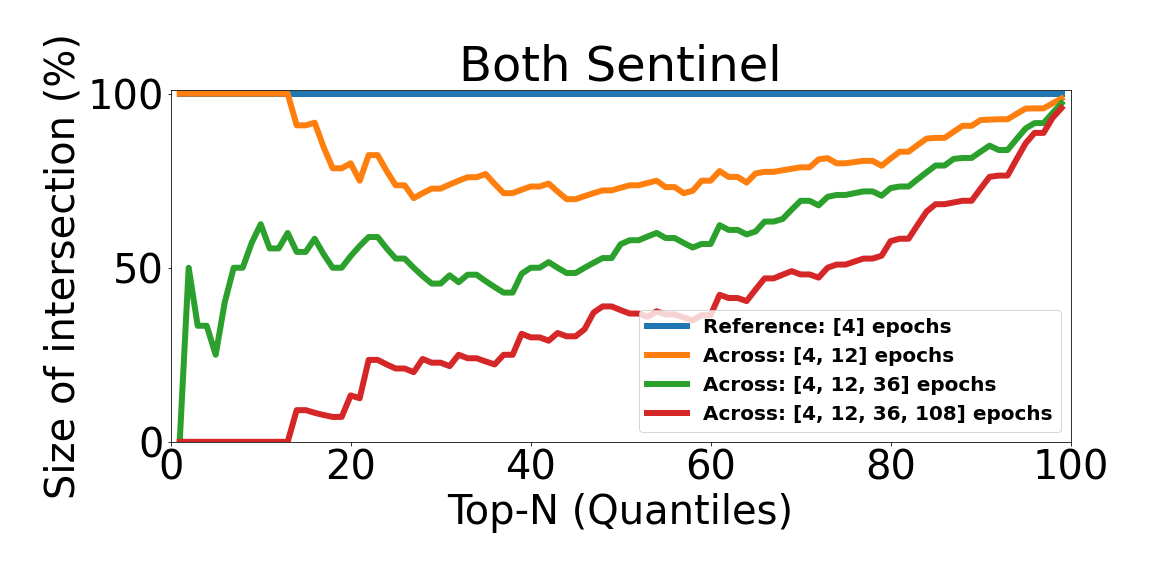}
    \caption{Positive persistence for Sentinel-1, Sentinel-2 and when using both sensors.} 
    \label{fig:persistence-positive}
\end{figure}

\begin{figure}[h]
\centering
    \includegraphics[width=0.4\textwidth]{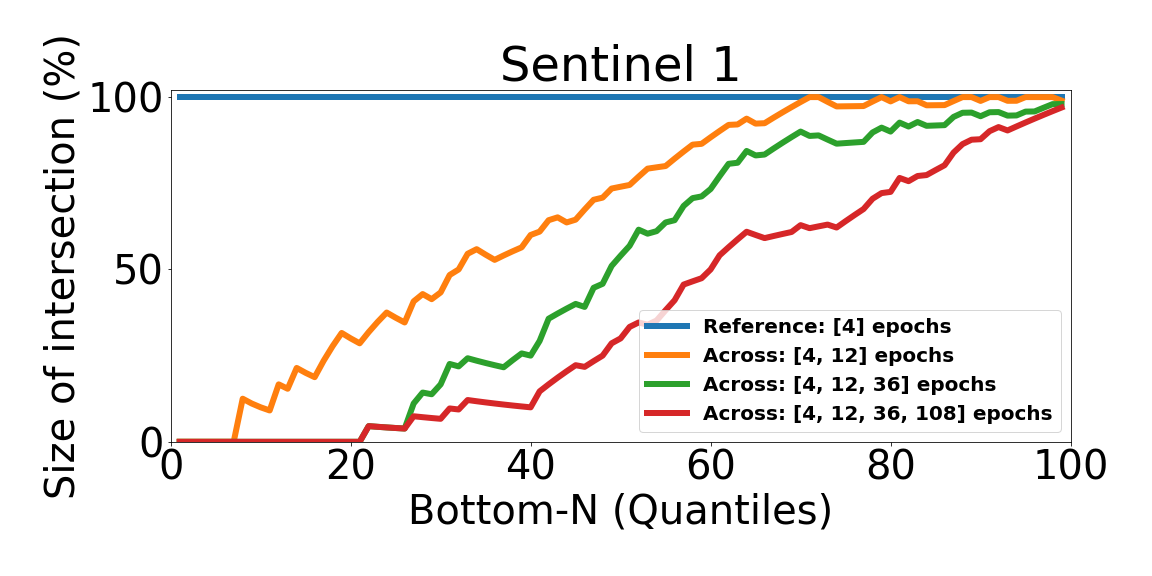}
    \qquad
    \includegraphics[width=0.4\textwidth]{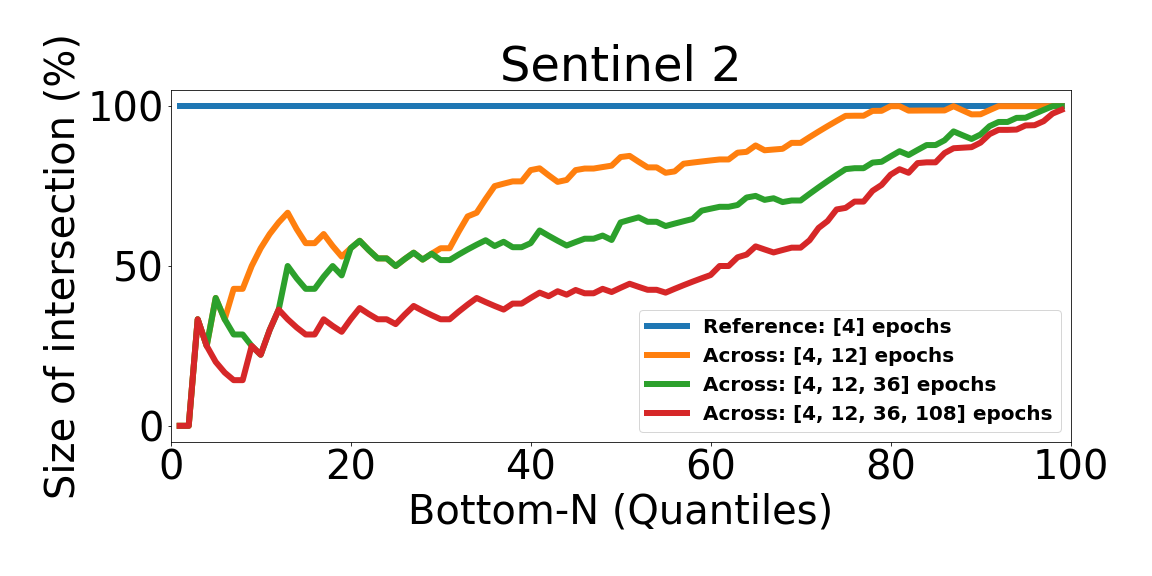}
     \qquad
    \includegraphics[width=0.4\textwidth]{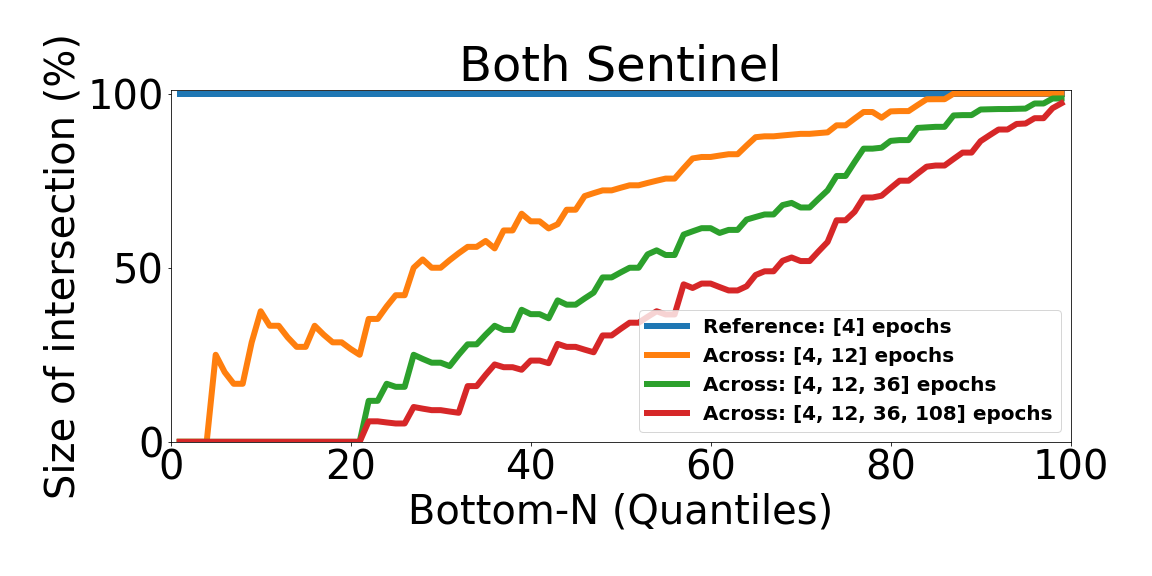}
    \caption{Negative persistence for Sentinel-1, Sentinel-2 and when using both sensors.} 
    \label{fig:persistence-negative}
\end{figure}

Figure~\ref{fig:persistence-positive} and \ref{fig:persistence-negative} 
show measurements of positive and negative persistence.
We consider samples collected for experiments related to section ~\ref{subsec:dos} and \ref{subsec:fdc}.
For each setting, the blue curve represents the reference population: the models at a given~$Rank-N$
based on their fitness after 4 epochs of training.
The yellow curve display the share of these models maintaining the same~$Rank-N$  after 12 epochs.
The green and red show the same (intersection of sets) respectively after 36 and 108 epochs of training.
The positive and negative Persistence refer to using the top and bottom ~$N$ rank function (Nth percentile).

First we have a look at the positive persistence (see Figure~\ref{fig:persistence-positive}).
Overall, we observe that the larger the fitness a sensor can provide (see Figure~\ref{fig:pdf-36} \ref{fig:pdf-108e}), 
the larger its  persistence (N$<$25) across all training budgets.
More precisely in terms of~\textit{Area under the Curve (N$<$25)}, Sentinel-2 ($AuC=0.14$) improves over the use of both sensors ($AuC=0.07$), which also improves over a single Sentinel-1 sensor ($AuC=0.04$).

Next we take a look at the negative persistence (see Figure~\ref{fig:persistence-negative}).
Overall, Sentinel-2 generates the larger persistence  ($P=31.82$, $AuC=0.26$), while the use of Sentinel-1 ($P=4.0$, $AuC=0.01$) or both sensors 
($P=5.26$, $AuC=0.01$) result in average measurements.
 
To summarize, we observe 
similar trend across sensors for both positive and negative persistence.
A larger fitting capacity results in a larger persistence (Sentinel-2).
In the case of Sentinel-1, the limited ability to fit the sensor 
might hinder the ability of models to keep their ranking (potential instabilities during training). 
In turn, this might result in a poorer persistence. 
For the better fitted sensor (Sentinel-2) chances of finding top-25\% and bottom-25\% performers are considerable ($P=13.7$, $P=31.82$).

\subsection{Fitness Landscape Footprint}\label{subsec:footprint}
The results obtained in the previous sections are summarized by the \textit{footprint} for each data source.
Figure~\ref{fig:footprnts} displays the \textit{footprint} for Sentinel-1 (blue), Sentinel-2 (yellow) and both sensors together (green).
This is done using considering only 36 epochs of training.

As observed in section~\ref{subsec:dos}, the search space appears better suited to fit Sentinel-2, than Sentinel-1. 
Indeed, Sentinel-2 enables reaching a larger mean fitness ($\mu=0.94$) and lower standard deviation ($\sigma=0.03$) than the other sensors.
Performing an input-level fusion slightly worsens the fitness ($\mu=0.89, \sigma=0.05$).
Similarly, the search landscape of Sentinel-2 appears more favourable to search trajectories. 
This transpires through random walk routes that are smoother and of higher fitness ($\mu=0.93$, $\tau=1.54$).
Regarding the evolution in the ranking of samples, 
a larger fitness results in a larger positive and negative persistence.
Indeed, this transpires in the larger measurements obtained for Sentinel-2 (\textit{Pos. AuC=0.14, Neg. AuC=0.26}).

\section{Conclusion}\label{sec:conclusion}

In this study, we investigate the impact of the choice of an input sensor 
on the performance of a neural architecture search strategy.
More precisely, we want to know to what extent does searching with a given sensor differs from searching with other sensors, 
in the context of neural architecture optimization.
Are there benefits or drawbacks in searching with fused sensors ?

To answer such questions, we use the framework of ~\textit{Fitness Landscape Footprint}, 
that we apply on the Real World image classification task So2Sat LCZ42, 
and analyze the process of searching CNN image classifiers provided in the NASBench-101 database.
After sampling and evaluating solutions with three different sensor settings (including input level fusion), 
we provide a comparative analysis assessing for instance the distribution of fitness, fitness distance correlation and ruggedness of the landscapes.

Overall, we observe a consistent improvement in the capacity of fitting all sensors, the longer the training time.
Using Sentinel-2 enables larger fitness, over Sentinel-1 or an input-level fusion of both sensors.
Similar results are observed when analysing search landscapes. Indeed, the longer the training, 
the landscape evolve from high ruggedness to flatness. 
Moreover, search strategies might benefit from a deployment on Sentinel-2, 
as it provides with routes that are smoother, of higher fitness, higher gain per distance travelled, 
and higher persistence in ranking of models.  
When a sensor can be fit well enough (Sentinel-2, fusion), we observe very similar behaviour in terms of trajectories (smoothness, ruggedness, fitness). 
This strongly indicates that search trajectories associated to different sensors are comparable 
when the search space is able to fit them decently enough. 

As future work, we propose to investigate how to use the gained insights to help build speed-up techniques for NAS strategies.
Such technique could rely, for instance, on searching with a sensor (or a subset of given sensors),
helping approximate the search with a more expensive to evaluate target sensor.

\section*{ACKNOWLEDGEMENTS}\label{ACKNOWLEDGEMENTS}
Authors acknowledge support by the European Research Council (ERC) under the European Union's Horizon 2020 research and innovation program (grant agreement No. [ERC-2016-StG-714087], Acronym: \textit{So2Sat}), by the Helmholtz Association
through the Framework of Helmholtz AI [grant  number:  ZT-I-PF-5-01] - Local Unit ``Munich Unit @Aeronautics, Space and Transport (MASTr)'' and Helmholtz Excellent Professorship ``Data Science in Earth Observation - Big Data Fusion for Urban Research'' (W2-W3-100),  by the German Federal Ministry of Education and Research (BMBF) in the framework of the international future AI lab "AI4EO -- Artificial Intelligence for Earth Observation: Reasoning, Uncertainties, Ethics and Beyond" (Grant number: 01DD20001) and the grant DeToL.

{
	\begin{spacing}{1.17}
		\normalsize
		\bibliography{bibliography} 
	\end{spacing}
}



\end{document}